\documentclass[a4paper,fleqn]{cas-sc}

\usepackage{times}
\usepackage{subfigure}
\usepackage{graphicx}
\usepackage{amsmath,amssymb}
\usepackage{bbold}
\usepackage{ragged2e}
\usepackage{multirow}
\usepackage{array}
\usepackage{url}
\usepackage{orcidlink}
\usepackage{algorithm}
\usepackage{algpseudocode}
\usepackage{rotating}
\usepackage{caption}
\usepackage[numbers]{natbib}

\graphicspath{{Figures/}}

\begin{document}
\let\WriteBookmarks\relax
\def\floatpagepagefraction{1}
\def\textpagefraction{.001}
\shorttitle{Mycelial Search for Continuous Optimisation}
\shortauthors{M. M. Dehshibi}

\title [mode = title]{Mycelial Search: A Graph-Structured Metaheuristic for Continuous Optimisation}                      
\author{Mohammad Mahdi {Dehshibi}~\orcidlink{0000-0001-8112-5419}}
\ead{mohammad.dehshibi@uwe.ac.uk}

\affiliation{
Unconventional Computing Laboratory, UWE Bristol, UK
}

\begin{abstract}
Continuous optimisation methods need to balance sharing information and maintaining alternative search directions. In this paper, we introduce Mycelial Search (Myco), a graph-structured metaheuristic designed around active tips, community-weighted flow, adaptive cord plasticity, and anchor-based injection. Candidate solutions form an evolving spatial graph in which a Louvain partition distinguishes within-community from cross-community information exchange. Adaptive cord plasticity subsequently modifies active tip-to-tip edges according to their alignment with the local flow. An anchor-based injection mechanism supplements the graph-driven tip dynamics. We evaluated Myco on the CEC 2022 single-objective bound-constrained benchmark suite at dimensions $D=10$ and $D=20$, using 30 independent runs per algorithm-function pair. The comparison includes eleven established optimisers from several search families. Myco reaches competitive results on selected functions across both dimensions. The ablation analysis further shows that community structure regulates the range of graph-based information exchange, whereas cord plasticity controls the persistence of local directional influence. These findings indicate that graph-structured local interaction can support continuous optimisation, while its effectiveness depends on landscape structure and information transfer across local search regions.
\end{abstract}

\begin{graphicalabstract}
    \includegraphics[width=\textwidth]{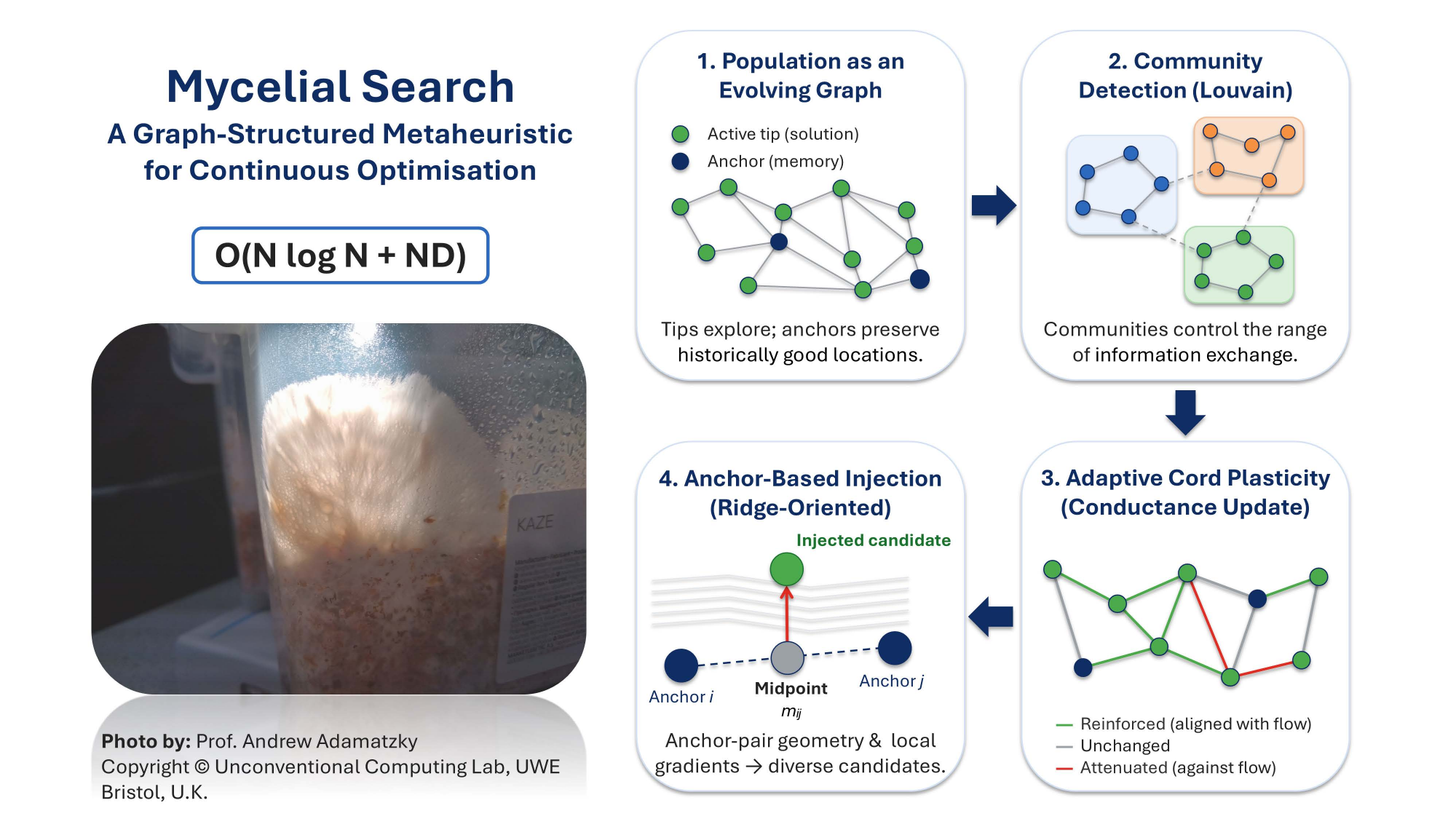}
\end{graphicalabstract}

\begin{highlights}
    \item Mycelial Search introduces an evolving graph structure for continuous optimisation.
    \item Community detection regulates information exchange across local search regions.
    \item Adaptive cord plasticity reinforces flow-aligned links and attenuates others.
    \item Per-iteration complexity scales as $O(N\log N+ND)$ under sparse graph connectivity.
    \item Myco achieves competitive results on CEC 2022 functions across two dimensions.
\end{highlights}


\maketitle

\section{Introduction}
\label{sec:introduction}

Continuous optimisation problems with interacting variables, multiple local optima, and regions with sharply different search behaviour appear in constrained engineering design and optimisation settings~\cite{dehshibi2017hybrid, machado2018revisiting, sepas2012novel, shabani2020search, turgut2023systematic}. Population-based metaheuristics address this setting by maintaining a set of candidate solutions and using their relative quality to guide the search. Genetic algorithms use selection and variation, particle swarm optimisation uses learned exemplar positions, and differential evolution constructs trial solutions from population differences~\cite{kennedy1995particle,storn1997differential,whitley1994genetic}. Later methods refined these principles through adaptive control parameters, external archives, and changing population sizes~\cite{brest2017single,tanabe2014improving,zhang2009jade}.

Many population methods still represent the relations among candidate solutions only weakly. Interaction is often mediated by a global best solution, sampled exemplars, or temporary population differences. These mechanisms can be effective, but they rarely retain a spatial interaction structure that changes with the search state. This matters when useful information is local. A candidate may benefit from nearby high-quality positions without requiring direct attraction to the same population-wide reference.

Graph-oriented optimisation offers a different perspective. Adaptive connectivity, flow, and edge reinforcement have been used to represent how information or resources move through a network~\cite{awad2023novel,gao2020accelerated}. Community detection identifies groups that are more densely connected internally than externally~\cite{blondel2008fast}. These approaches commonly address problems in which a graph, route, or transport network is itself the object of optimisation. Candidate solutions in continuous search can be organised the same way as nodes in an evolving graph, where local connectivity determines which search information is exchanged and how strongly it contributes to subsequent motion.

Fungal mycelia offer a suitable biological motivation for this perspective. A mycelial network coordinates distributed exploration with transport across the colony. Established pathways can remain available while local connectivity adapts to usage patterns~\cite{adamatzky2022fungal, dehshibi2021electrical, dehshibi2023complexity, dehshibi2021stimulating, dehshibi2023stimulating}. We use this principle without attempting to reproduce fungal physiology. We introduce Mycelial Search (Myco), a graph-structured metaheuristic for continuous optimisation. Myco represents candidate solutions as active tips and retains a bounded set of anchors at historically favourable positions within a spatial graph that is reconstructed at every iteration. Tips receive locally weighted flow information from neighbouring nodes. Anchors preserve high-quality locations that remain accessible to the evolving graph.

Myco combines two forms of graph organisation. A Louvain partition imposes a mesoscale organisation on the current graph and distinguishes within-community from cross-community interactions~\cite{blondel2008fast}. A conductance update then modifies the current tip-to-tip edges based on their directional agreement with the induced flow, reinforcing flow-aligned edges and attenuating unsupported ones. The community partition regulates the scope of information exchange across the graph. The conductance update retains short-term directional information within the local graph.

We evaluate Myco on the CEC 2022~\cite{cec2022} single-objective bound-constrained benchmark suite across two problem dimensions and compare it with established optimisation methods from several search families. We also examine the separate contributions of adaptive conductance and community partitioning. This analysis tests whether an evolving local interaction graph and adaptive edge strength provide useful search information in continuous optimisation.

The rest of this paper is organised as follows: Section~\ref{sec:related_work} reviews population-based and graph-oriented optimisation. Section~\ref{sec:proposed_method} presents Myco and its graph construction, community-weighted flow, adaptive conductance, and auxiliary search components. Section~\ref{sec:experiments} describes the experimental protocol, comparative results, and ablation analyses. The final section concludes the paper and identifies limitations of the current design.

\section{Related Work}
\label{sec:related_work}

Research on metaheuristic optimisation has produced a wide range of algorithms that may appear distinct at the metaphorical level, yet often share similar underlying computational structures~\cite{ma2023performance, nanda2014survey}. In many approaches, candidate solutions are updated using individual-level rules and exchange information through limited elite references or population-wide guidance, even as operator design and search configuration become more sophisticated~\cite{wu2019ensemble}. Although such designs have led to many effective optimisers, they often provide only limited support for local cooperation, spatial grouping, and short-term structural memory. To address these limitations, research studies have increasingly explored methods that explicitly incorporate agent connectivity and population organisation into the search process, rather than relying solely on flat population-level updates~\cite{ma2019multi}.

\subsection{Population-Based Search}
\label{sec:rw-population-families}

Population-based optimisation is based on the idea that a set of concurrently updated candidate solutions can balance exploration and exploitation more effectively than a purely local, single-solution procedure. Genetic algorithm (GA) relies on population evolution through selection and variation~\cite{whitley1994genetic}. Particle swarm optimisation (PSO) updates each particle by combining its own search history with information obtained from exemplar positions in the swarm~\cite{kennedy1995particle}. Differential evolution (DE) generates trial solutions by taking vector differences among population members and has become a foundation of continuous global optimisation~\cite{storn1997differential}.

Subsequent work focused on increasing search quality by refining how individuals learn, mutate, or adapt their control parameters. CLPSO widened the learning source of each particle and improved diversity maintenance in multimodal search through comprehensive exemplar selection~\cite{liang2006comprehensive}. In the DE family, SAP-DE explored self-adaptive population sizing within differential evolution~\cite{teo2005differential}, JADE introduced adaptive differential evolution with an optional archive~\cite{zhang2009jade}, SHADE formalised success-history-based parameter adaptation~\cite{tanabe2013success}, L-SHADE added linear population size reduction~\cite{tanabe2014improving}, and jSO continued this progression toward high-performance real-parameter optimisation~\cite{brest2017single}. While these methods show that substantial gains can be obtained by improving parameter control, mutation design, and learning strategy within a shared population-update framework, their search logic remains centred on updating individuals within a flat population.

Bio-inspired optimisers reformulated the population search logic by emulating various natural processes. Artificial bee colony (ABC) modelled foraging behaviour through employed, onlooker, and scout bees~\cite{karaboga2006powerful}. The grey wolf optimiser (GWO) and the whale optimisation algorithm (WOA) describe the search process through leadership hierarchy and hunting behaviour~\cite{mirjalili2016whale,mirjalili2014grey}. The slime mould algorithm (SMA) translates oscillatory adaptation in slime mould into a stochastic optimiser~\cite{li2020slime}. The moss growth optimisation (MGO) is a recent example that organises the search through wind-direction estimation, spore dispersal, dual propagation, and cryptobiosis~\cite{zheng2024moss}. Despite their differences in inspiration and update mechanisms, these algorithms still update a population whose relational structure is only weakly modelled. This observation is important for the present paper because it motivates the shift from a flat population of individually updated agents to graph-structured search, in which candidate solutions interact through an explicitly defined and evolving interaction graph.

\subsection{Network- and Graph-Oriented Optimisation}
\label{sec:rw-network-optimisation}

A more structured alternative to flat population search appears in optimisation methods that operate through adaptive graphs. Instead of relying solely on isolated update rules, these approaches allow connectivity, flow balance, and path reinforcement to influence the evolution of the search over time~\cite{awad2023novel}. Physarum-inspired optimisation is a representative case, since it converts adaptive transport behaviour into graph-based computational dynamics~\cite{gao2020accelerated}.

Such methods have mainly been developed for network-centred problems. The accelerated Physarum solver demonstrated that the number of inactive nodes can be reduced and computation can be terminated earlier without sacrificing effectiveness in network optimisation tasks~\cite{gao2020accelerated}. Physarum-inspired routing methods have similarly used adaptive graph dynamics to reconstruct favourable transmission paths in changing communication environments~\cite{martinelli2025bioinspired}. 

While these studies show that evolving network structure can guide optimisation, they do not fully address the setting considered in this study. Graph-oriented methods were designed for problems where the solution is already a path, route, or transport network~\cite{awad2023novel, martinelli2025bioinspired}. Mycelial Search (Myco) is motivated by this gap. It adopts a graph-based perspective on adaptive connectivity and flow-based reinforcement to organise interactions among candidate solutions in continuous optimisation. Therefore, the evolving network becomes a search mechanism rather than only a problem-specific representation.

\section{Proposed Method}
\label{sec:proposed_method}

This section presents the Mycelial Search (Myco) at three connected levels. We first state the biological motivation that guided the design. We then define the network construction, the community-driven flow mechanism, the cord plasticity rule, and the anchor and Ridge-Oriented Injection (ROI) updates. Finally, we summarise the full algorithm and its computational cost.

\subsection{Biological Motivation}
\label{sec:biological_motivation}

The proposed method is motivated by the way fungal mycelia coordinate exploration and transport through a distributed network~\cite{adamatzky2022fungal}. Fungal mycelia solve two linked problems at once: expanding into new regions while maintaining transport across the already-formed network. This dual behaviour motivates our proposed Mycelial Search, which formulates search as a spatial network in which candidate solutions interact through local connections, directional flow, and adaptive edge strength.

Myco does not model chemotropic growth or substrate-level physiology. Instead, it borrows the idea of many local explorers moving concurrently through a shared networked search space. Tips represent active exploratory units. Anchors preserve historically strong locations and stabilise the evolving network. Together, these components reflect three recurring features of mycelial organisation: exploration remains distributed, promising locations remain accessible to the rest of the network, and information is exchanged through local neighbourhoods rather than a single global communication pattern.

The strongest biological link in the method is the conductance update on tip-to-tip connections. In fungal transport networks, frequently used pathways tend to persist, whereas weakly used pathways regress~\cite{adamatzky2023reactive, dehshibi2021electrical, dehshibi2021stimulating}. The proposed method implements the same directional idea through adaptive cord conductance. Edges aligned with flow are reinforced, while edges not supported by flow decay over time. This mechanism is central to the method, where the community structure enters at a different level. The Louvain partition~\cite{blondel2008fast} introduces an intermediate scale of organisation. It is biologically motivated only in the limited sense that real mycelial networks exhibit spatially organised domains. However, the partition itself is algorithmic rather than biological. ROI remains outside the main biological claim and is included only as an auxiliary search mechanism.

\subsection{Network Construction}
\label{sec:network_construction}

We represent the current search state as an undirected, weighted spatial graph $\mathcal{G}$ whose nodes consist of active tips and retained anchors. Let $D$ denote the problem dimension, $N$ the number of tips, and $K$ the number of retained anchors, where $K \leq K_{\max}$. The position of tip $i$ is denoted by $x_{i} \in \mathbb{R}^{D}$ for $i=1,\dots,N$, and the position of anchor $k$ is denoted by $a_{k} \in \mathbb{R}^{D}$ for $k=1,\dots,K$. The corresponding node set is $\mathcal{V} = \{x_{1},\dots,x_{N},a_{1},\dots,a_{K}\}$.

To keep the retained-anchor set compact, we cap the set at $K_{\max}$. This way, if more than $K_{\max}$ anchors are available, only the $K_{\max}$ anchors with the lowest objective values are retained.

Connectivity is induced locally through a fixed fusion radius. Let $\mathbf{l} \in \mathbb{R}^{D}$ and $\mathbf{u} \in \mathbb{R}^{D}$ denote the lower and upper bounds of the search space. We set the radius to $r_{\mathrm{fuse}} = 0.1 \left\| \mathbf{u} - \mathbf{l} \right\|_2$. The radius criterion determines only whether an edge is present. Since it does not specify the interaction strength between connected nodes, we quantify the effective edge weight using Eq.~\eqref{eq:edge_weight}.

\begin{equation}
    e_{pq} = \frac{g_{pq}}{\|z_{p}-z_{q}\|_{2} + \varepsilon}, \quad \varepsilon = 10^{-6}
    \label{eq:edge_weight}
\end{equation}
where $g_{pq}$ denotes the stored conductance associated with the undirected pair $(p,q)$, and $z_{p}$ and $z_{q}$ are the position of nodes $p$ and $q$ in $\mathcal{V}$

The conductance is shared symmetrically by $(p,q)$ and $(q,p)$. Previously unseen tip-to-tip pairs are initialised with $g_{pq}=1$, whereas anchor-incident edges retain fixed conductance values and are not modified by plasticity. Pairs that fail the radius condition are simply absent from the graph at that iteration. In this construction, anchors serve as elite landmarks that shape the local neighbourhood structure and influence the subsequent flow field without themselves becoming adaptive cords. We restrict conductance adaptation to tip-to-tip edges so that anchors provide stable reference points while the search network remains free to reorganise around them.

\subsection{Cord Plasticity}
\label{sec:cord_plasticity}

We use cord plasticity to provide a short-term memory mechanism on the weighted graph by reinforcing directions that are repeatedly supported by the induced flow. The rule applies only to present tip-to-tip edges in the current graph. Let $\mathcal{N}_{i}$ denote the set of all neighbours of tip $i$ in the current graph, including anchors when present, and let  $\mathcal{N}_{i}^{\mathrm{tip}} \subseteq \mathcal{N}_{i}$ denote the subset containing only tip neighbours. For each undirected tip pair $(i,j)$ with $j \in \mathcal{N}_{i}^{\mathrm{tip}}$, the conductance is stored as a single shared scalar $g_{ij}=g_{ji}$. Edges incident to anchors remain part of the graph and contribute to the flow field, but they are excluded from conductance adaptation. Let $\phi_{i} \in \mathbb{R}^D$ be the flow vector associated with tip $i$, and $\tau_{0}$ denote a numerical tolerance. For $\|\phi_i\|_2 \geq \tau_0$, we normalize the local flow according to Eq.~\eqref{eq:plasticity_unit_flow}.

\begin{equation}
    \hat{\phi}_i=\frac{\phi_i}{\|\phi_i\|_2}.
    \label{eq:plasticity_unit_flow}
\end{equation}

To define the edge direction vector, we use $c_{ij}=x_j-x_i$ for each $j \in \mathcal{N}_i^{\mathrm{tip}}$. If $\|c_{ij}\|_2 < \tau_0$, the edge is left unchanged during the update of tip $i$. Otherwise, directional agreement between the edge and the local flow is measured by a cosine alignment score defined in Eq.~\eqref{eq:plasticity_alignment}.

\begin{equation}
    s_{ij}=\frac{c_{ij}^{\top}\hat{\phi}_i}{\|c_{ij}\|_2}.
    \label{eq:plasticity_alignment}
\end{equation}

Given the alignment score in Eq.~\eqref{eq:plasticity_alignment}, reinforcement is applied only when $s_{ij}>\tau_{\mathrm{align}}$, where $\tau_{\mathrm{align}}$ is the alignment threshold. In that case, the reinforced conductance is obtained from Eq.~\eqref{eq:plasticity_reinforcement}.

\begin{equation}
    \tilde{g}_{ij}=\min\!\left(g_{\max},\,g_{ij}(1+\eta s_{ij})\right),
    \label{eq:plasticity_reinforcement}
\end{equation}
where $\eta$ is the reinforcement rate and $g_{\max}$ is the upper conductance bound. For $s_{ij}\le\tau_{\mathrm{align}}$, we set $\tilde{g}_{ij}=g_{ij}$. Decay with lower clipping is then applied through Eq.~\eqref{eq:plasticity_decay}.

\begin{equation}
    g_{ij}^{+}=\max\!\left(g_{\min},\,(1-\delta)\tilde{g}_{ij}\right).
    \label{eq:plasticity_decay}
\end{equation}
where $\delta$ is the decay rate and $g_{\min}$ is the lower conductance bound.

Because plasticity is evaluated once from the perspective of each tip, the same undirected conductance may be revised twice within a single iteration, once when processing endpoint $i$ and once when processing endpoint $j$. Therefore, the final end-of-iteration value may include a second update from the opposite endpoint. This update structure preserves conductance symmetry because both endpoints modify the same shared scalar. As a result, edges repeatedly aligned with the induced flow are first reinforced using Eq.~\eqref{eq:plasticity_reinforcement} and then decayed and clipped using Eq.~\eqref{eq:plasticity_decay}, whereas in the low-flow regime only Eq.~\eqref{eq:plasticity_decay} is applied.

\subsection{Community Structure and Mesoscale Flow}
\label{sec:community_flow}

We use the Louvain method~\cite{blondel2008fast} to partition $\mathcal{G}$ into communities at each iteration. For all reported experiments, we fix the community-detection random seed to 42. The raw partition is refined by a two-criterion filter. A community is retained if it contains at least $n_{\min}$ nodes and its induced subgraph density satisfies $\rho \geq \rho_{\min}$, where $\rho = 2m/(n(n-1))$ for a subgraph with $n$ nodes and $m$ internal edges. Nodes belonging to rejected communities are each reassigned a unique singleton label. They remain in the graph with all edges preserved and continue to contribute to the flow computation on equal terms with all other nodes. After reassignment, interactions involving singleton-labelled nodes use $\lambda_{\mathrm{inter}}$ in Eq.~\eqref{eq:message_weight}.

We define a potential for every node $v \in \mathcal{V}$, over both tips and anchors. Let $f(v)$ denote the objective value associated with node $v$. With $f_{\min}$ and $f_{\max}$ computed over all nodes in $\mathcal{V}$ at the current iteration, the potential is defined in Eq.~\eqref{eq:potential}.

\begin{equation}
    s(v) = \left\{ 
    \begin{array}{cl}
        \frac{f_{\max} \, - f(v)}{f_{\max} \, - f_{\min} \, + \, \varepsilon}, & f_{\max} \neq f_{\min} \\
        1, & \mathrm{otherwise}
    \end{array} \right., \qquad \varepsilon = 10^{-6}
    \label{eq:potential}
\end{equation}

For tip $i$, the flow vector $\phi_{i}$ is defined as a weighted mean of spatial displacements toward all neighbours $j \in \mathcal{N}_{i}$ in the current graph, including anchors. We define the message weight from node $j$ to tip $i$ as in Eq.~\eqref{eq:message_weight}.

\begin{equation}
    m_{ij} = \lambda(i,j) \cdot e_{ij} \cdot \sigma_{ij},
    \label{eq:message_weight}
\end{equation}
where $e_{ij}$ is the edge weight. The community factor $\lambda(i,j)$ takes the value $\lambda_{\mathrm{intra}}$ for node pairs in the same filtered community and $\lambda_{\mathrm{inter}}$ for node pairs in different filtered communities, where $\lambda_{\mathrm{intra}}$ and $\lambda_{\mathrm{inter}}$ are constant weights for within-community and cross-community interactions, respectively. The sigmoid gate is defined in Eq.~\eqref{eq:sigmoid_gate}.

\begin{equation}
    \sigma_{ij} = \frac{1}{1 + \exp\!\left(-\kappa\,(s_{j} - s_{i})\right)}, 
    \label{eq:sigmoid_gate}
\end{equation}
where $\sigma_{ij}$ modulates the contribution according to the potential difference between neighbouring node $j$ and receiving tip $i$. We compute the flow vector $\phi_{i}$ as the weighted mean of spatial displacements toward the neighbours of tip $i$ using Eq.~\eqref{eq:flow_vector}.

\begin{equation}
    \phi_{i} = \frac{\sum_{j \in \mathcal{N}_{i}} m_{ij}\,(z_{j} - z_{i})}{\sum_{j \in \mathcal{N}_{i}} m_{ij}},
    \label{eq:flow_vector}
\end{equation}
where $z_i$ and $z_j$ are the position of node $i$ and $j$ in $\mathcal{V}$, respectively. When the total message weight is zero, $\phi_{i} = \mathbf{0}$. We anneal the sigmoid steepness $\kappa$ linearly over the run. Let $\mathrm{FE}(t)$ denote the cumulative number of function evaluations consumed up to the start of iteration $t$, and let $\mathrm{FE}_{\max}$ be the total evaluation budget. Then Eq.~\eqref{eq:kappa_anneal} defines the linear annealing schedule for $\kappa$.

\begin{equation}
    \kappa(t) = \kappa_{\max} - (\kappa_{\max} - \kappa_{\min})\,\frac{\mathrm{FE}(t)}{\mathrm{FE}_{\max}}.
    \label{eq:kappa_anneal}
\end{equation}

A larger $\kappa$ makes the sigmoid more selective with respect to potential differences, suppressing contributions from neighbours with lower potential. As $\kappa$ decreases toward $\kappa_{\min}$, the gate becomes less selective and the flow aggregates more uniformly across the neighbourhood.

\subsection{Tip Update and Anchor Management}
\label{sec:tip_update_anchor}

We update the tip states and the retained-anchor set within the same iteration. We first evaluate the tips and update the incumbent best position $x^\star(t)$, which is the only position eligible for anchor insertion. We insert this position only when its Euclidean distance from every retained anchor exceeds $\varepsilon = 10^{-6}$. If the number of retained anchors exceeds $K_{\max}$, we keep the anchors with the best stored objective values and reevaluate the retained anchors before computing the node-wise potentials. These updated anchor values are then used in the current potential field and the resulting flow computation. When ROI is active, we also use values to evaluate the relative fitness gap that triggers the mechanism.

Let $v_{i}^{t} \in \mathbb{R}^{D}$ denote the velocity of tip $i$ at iteration $t$ and $r_{i}^{t} \in [0,1]^{D}$ be a random vector sampled independently for tip $i$. We perturb the current velocity of tips with at most one graph neighbour before applying the main velocity update. This perturbation is defined in Eq.~\eqref{eq:isolated_kick}.

\begin{equation}
    \tilde{v}_{i}^{t} =
    \begin{cases}
        v_i^t + \xi_i^t, & |\mathcal{N}_i| \leq 1 \\
        v_i^t, & |\mathcal{N}_i| > 1
    \end{cases},
    \quad
    \xi_i^t \sim \mathcal{U}([-\delta,\delta]^D),
    \quad
    \delta = 0.05\,r_{\mathrm{fuse}},
    \label{eq:isolated_kick}
\end{equation}
where $\mathcal{U}([-\delta,\delta]^D)$ denotes the uniform distribution over the hypercube $[-\delta,\delta]^D$. We then calculate the velocity of tip $i$ at iteration $t+1$ using Eq.~\eqref{eq:velocity_update}.

\begin{equation}
    v_{i}^{t+1} = w\,\tilde{v}_{i}^{t} + c_{\mathrm{exp}}\,r_{i}^{t} \odot (x^\star(t) - x_{i}^{t}) + c_{\mathrm{flow}}\,\phi_{i},
    \label{eq:velocity_update}
\end{equation}
where $w$ is the inertia coefficient, $c_{\mathrm{exp}}$ is the coefficient of the best-position term, $c_{\mathrm{flow}}$ is the coefficient of the flow term, and $\odot$ denotes element-wise multiplication. We update the tip position through Eq.~\eqref{eq:tip_position_update}.

\begin{equation}
    x_i^{t+1} = x_i^t + v_i^{t+1}.
    \label{eq:tip_position_update}
\end{equation}

We clip each component of $x_i^{t+1}$ to the corresponding interval defined by $\mathbf{l}$ and $\mathbf{u}$ after the position update.

\subsection{Ridge-Oriented Injection Spawning}
\label{sec:roi_spawning}

Ridge-Oriented Injection (ROI) is an auxiliary mechanism that injects a candidate position derived from the retained-anchor set. It is considered only when at least two anchors are present. We sample two distinct anchors $a_{k_{1}}$ and $a_{k_{2}}$ uniformly, without replacement, and evaluate their relative fitness gap using Eq.~\eqref{eq:roi_gap}.

\begin{equation}
    \gamma =
    \frac{\lvert f(a_{k_{1}}) - f(a_{k_{2}}) \rvert}
         {\max_{1 \leq k \leq K} f(a_{k}) \,-\, \min_{1 \leq k \leq K} f(a_{k}) \,+\, \varepsilon}.
    \label{eq:roi_gap}
\end{equation}

The ROI branch proceeds when $\gamma > \tau_{\mathrm{roi}}$, where $\tau_{\mathrm{roi}}$ is the relative-gap threshold. The midpoint $x_{\mathrm{mid}} = (a_{k_1} + a_{k_2})/2$ serves as the base for the spawn construction. We estimate $\nabla f(x_{\mathrm{mid}})$ by central finite differences, as specified in Eq.~\eqref{eq:roi_gradient}.

\begin{equation}
    \bigl[\nabla f(x_{\mathrm{mid}})\bigr]_d =
    \frac{f(x_{\mathrm{mid}} + h\,\mathbf{e}_d) - f(x_{\mathrm{mid}} - h\,\mathbf{e}_d)}{2h},
    \quad d = 1,\dots,D,
    \label{eq:roi_gradient}
\end{equation}
where $\mathbf{e}_d$ denotes the $d$-th standard basis vector and $h = 10^{-5}\max\{\lvert f(x_{\mathrm{mid}}) \rvert, 1\}$ is the step size. In addition to the evaluation of $f(x_{\mathrm{mid}})$ needed to set $h$, this sub-step requires $2D + 1$ objective evaluations.

The geometric adjustment preserves only the gradient component orthogonal to the anchor axis $u = (a_{k_2} - a_{k_1})\,/\,\lVert a_{k_2} - a_{k_1} \rVert_2$, and the transverse component is derived using Eq.~\eqref{eq:roi_perp}.

\begin{align}
    \label{eq:roi_perp}
    p_\perp & = \hat{p} - (\hat{p}^\top u)\,u,\\ \nonumber
    \hat{p} & = \frac{\nabla f(x_{\mathrm{mid}})}{\lVert \nabla f(x_{\mathrm{mid}}) \rVert_2},
\end{align}

When either $\lVert a_{k_2} - a_{k_1} \rVert_2$ or $\lVert \nabla f(x_{\mathrm{mid}}) \rVert_2$ is below $\tau_{\textrm{deg}} = 10^{-9}$, the correction degenerates and $x_{\mathrm{mid}}$ is passed through unchanged. The spawn candidate is then calculated using Eq.~\eqref{eq:roi_spawn}.
\begin{equation}
    x_{\mathrm{roi}} = \Pi_{[\mathbf{l},\mathbf{u}]}\!\left(x_{\mathrm{mid}} - \alpha_{\mathrm{roi}}\,p_\perp\right),
    \label{eq:roi_spawn}
\end{equation}
where $\Pi_{[\mathbf{l},\mathbf{u}]}$ denotes component-wise clipping to the feasible domain and $\alpha_{\mathrm{roi}}$ is the ROI correction coefficient. We evaluate $x_{\mathrm{roi}}$ and replace the current worst tip only when the spawned value is strictly lower, adding one objective evaluation to the cost of the triggered ROI branch.

\subsection{Algorithm Summary and Complexity}
\label{sec:algorithm_complexity}

Algorithm~\ref{alg:proposed_method} summarises the complete procedure. We use it to consolidate the interaction between the retained-anchor mechanism, the community-weighted flow field, the conductance update, and the tip dynamics defined in Section~\ref{sec:network_construction} through Section~\ref{sec:roi_spawning}. The resulting iteration uses the current tip and anchor evaluations to form the potential field, applies the filtered Louvain partition to the flow computation, updates tip-tip conductances based on the induced flow, and then advances the tips. When ROI is enabled, the same iteration may also inject one additional candidate derived from a selected anchor pair.

\par\smallskip
\noindent\rule{\columnwidth}{0.5pt}\par
\refstepcounter{algorithm}
\noindent\textbf{Algorithm \thealgorithm. Mycelial Search}
\label{alg:proposed_method}\vspace{-0.5\baselineskip}\par
\noindent\rule{\columnwidth}{0.5pt}\par
\smallskip

\begin{algorithmic}[1]
\Require Objective function $f$, dimension $D$, bounds $\mathbf{l},\mathbf{u}$, number of tips $N$, anchor cap $K_{\max}$, evaluation budget $\mathrm{FE}_{\max}$, model hyperparameters
\Ensure Incumbent best position $x^{\star}(t)$ and, when needed, its objective value $f^{\star}(t)$

\Statex \vspace{-0.3\baselineskip}
\Statex \texttt{\% Initialization \%}

\State $r_{\mathrm{fuse}} \gets 0.1\|\mathbf{u}-\mathbf{l}\|_2$
\State Initialize $x_i \sim \mathcal{U}([\mathbf{l},\mathbf{u}])$ and $v_i^0 \gets \mathbf{0}$ for $i=1,\dots,N$
\State Initialize retained anchors $\mathcal{A}\gets\emptyset$ and the conductance map for tip-tip pairs
\State Set $\mathrm{FE}(t)\gets 0$, $f^{\star}(t)\gets +\infty$, and $x^{\star}(t)\gets \mathrm{null}$

\While{$\mathrm{FE}(t) < \mathrm{FE}_{\max}$}

    \Statex \vspace{-0.3\baselineskip}
    \Statex \texttt{\% Annealed Sigmoid Steepness \%}
    \State Update $\kappa(t)$ using Eq.~\eqref{eq:kappa_anneal} with the cumulative evaluation count at the start of the current iteration

    \Statex \vspace{-0.3\baselineskip}
    \Statex \texttt{\% Tip Evaluation and Incumbent Update \%}
    \State Evaluate all current tips and increment $\mathrm{FE}(t)$ by $N$
    \State Update $f^{\star}(t)$ and $x^{\star}(t)$ if a better tip is found

    \Statex \vspace{-0.3\baselineskip}
    \Statex \texttt{\% Anchor Management \%}
    \State Insert $x^{\star}(t)$ into $\mathcal{A}$ only if its Euclidean distance from every retained anchor exceeds $10^{-6}$
    \If{$|\mathcal{A}| > K_{\max}$}
        \State Retain the $K_{\max}$ anchors with the best stored objective values
    \EndIf
    \State Reevaluate all retained anchors and increment $\mathrm{FE}(t)$ by $|\mathcal{A}|$

    \Statex \vspace{-0.3\baselineskip}
    \Statex \texttt{\% Node-wise Potentials from Already Available Objective Values \%}
    \State Form the node set $\mathcal{V}=\{x_1,\dots,x_N\}\cup\mathcal{A}$
    \State Collect the already evaluated objective values $\{f(v): v\in\mathcal{V}\}$ from the current tips and retained anchors
    \State Compute $s(v)$ for all $v\in\mathcal{V}$ using Eq.~\eqref{eq:potential}

    \Statex \vspace{-0.3\baselineskip}
    \Statex \texttt{\% Graph Construction and Community Structure \%}
    \State Construct the current graph $\mathcal{G}$ using $r_{\mathrm{fuse}}$
    \State Compute edge weights $e_{pq}$ using Eq.~\eqref{eq:edge_weight}
    \State Compute the Louvain partition of $\mathcal{G}$ with seed $42$
    \State Filter the raw communities using $n_{\min}$ and $\rho_{\min}$, then relabel rejected communities as singletons

    \Statex \vspace{-0.3\baselineskip}
    \Statex \texttt{\% Mesoscale Flow \%}
    \For{$i\gets 1$ to $N$}
        \State Compute $\phi_i$ from Eqs.~\eqref{eq:message_weight}--\eqref{eq:flow_vector}
        \Statex \hspace{\algorithmicindent}over all $j\in\mathcal{N}_i$, using the previously computed
        \Statex \hspace{\algorithmicindent}potentials and filtered community labels
    \EndFor

    \Statex \vspace{-0.3\baselineskip}
    \Statex \texttt{\% Cord Plasticity \%}
    \For{$i\gets 1$ to $N$}
        \State Update the conductances of present tip-tip edges $(i,j)$ with
        \Statex \hspace{\algorithmicindent}$j\in\mathcal{N}_i^{\mathrm{tip}}$ using the previously computed
        \Statex \hspace{\algorithmicindent}flow vector $\phi_i$ and
        \Statex \hspace{\algorithmicindent}Eqs.~\eqref{eq:plasticity_unit_flow}--\eqref{eq:plasticity_decay}
    \EndFor

    \Statex \vspace{-0.3\baselineskip}
    \Statex \texttt{\% Tip Motion \%}
    \For{$i\gets 1$ to $N$}
        \If{$|\mathcal{N}_i|\leq 1$}
            \State Sample $\xi_i^t \sim \mathcal{U}([-\delta,\delta]^D)$ with $\delta=0.05\,r_{\mathrm{fuse}}$
            \State Set $\tilde{v}_i^t \gets v_i^t + \xi_i^t$
        \Else
            \State Set $\tilde{v}_i^t \gets v_i^t$
        \EndIf
        \State Sample $r_i^t \sim \mathcal{U}([0,1]^D)$
        \State Update $v_i^{t+1}$ using Eq.~\eqref{eq:velocity_update}
        \State Update $x_i^{t+1}$ using Eq.~\eqref{eq:tip_position_update}
        \State Clip $x_i^{t+1}$ component-wise to $[\mathbf{l},\mathbf{u}]$
    \EndFor

    \Statex \vspace{-0.3\baselineskip}
    \Statex \texttt{\% Auxiliary ROI Branch \%}
    \If{ROI is enabled and $|\mathcal{A}|\geq 2$}
        \State Sample two distinct anchors $a_{k_1},a_{k_2}\in\mathcal{A}$ uniformly without replacement
        \State Compute the relative gap $\gamma$ using Eq.~\eqref{eq:roi_gap}
        \If{$\gamma > \tau_{\mathrm{roi}}$}
            \State Set $x_{\mathrm{mid}} \gets (a_{k_1}+a_{k_2})/2$
            \State Estimate $\nabla f(x_{\mathrm{mid}})$ using Eq.~\eqref{eq:roi_gradient}
            \State Update $\mathrm{FE}(t)\gets \mathrm{FE}(t)+2D+1$
            \State Compute $p_\perp$ using Eq.~\eqref{eq:roi_perp}
            \State Construct $x_{\mathrm{roi}}$ using Eq.~\eqref{eq:roi_spawn}
            \State Evaluate $x_{\mathrm{roi}}$ and update $\mathrm{FE}(t)\gets \mathrm{FE}(t)+1$
            \State Replace the current worst tip only if $x_{\mathrm{roi}}$ yields a strictly lower objective value
        \EndIf
    \EndIf

\EndWhile

\State \Return $x^{\star}(t)$ and $f^{\star}(t)$
\end{algorithmic}
\par\noindent\rule{\columnwidth}{0.5pt}
\par\medskip

Let $K \leq K_{\max}$ denote the number of retained anchors at the current iteration, let $M = N + K$ denote the total number of graph nodes, and let $E = |\mathcal{E}|$ denote the number of graph edges. Excluding the cost of objective-function evaluation, graph construction with spatial neighbourhood queries requires $O(M\log M + E)$ operations. The mesoscale flow computation and the cord-plasticity update each scale as $O(ED)$. The velocity and position updates scale as $O(ND)$. Therefore, the per-iteration arithmetic cost is given by Eq.~\eqref{eq:iter_complexity}.

\begin{equation}
    O(M\log M + ED + ND).
    \label{eq:iter_complexity}
\end{equation}

We cap the anchor set by $K_{\max}$, and the fusion radius induces local connectivity. In the sparse regime considered here, $E$ therefore grows approximately linearly with $M$. Under this condition, Eq.~\eqref{eq:iter_complexity} reduces in practice to $O(N\log N + ND)$. Each iteration requires $N + K$ objective evaluations from tip evaluation and anchor reevaluation. When the ROI branch is triggered, it adds $2D+2$ additional evaluations: $2D+1$ for the midpoint-based numerical gradient and 1 for the spawned candidate. Since $K \leq K_{\max}$, the baseline evaluation cost per iteration remains linear in $N$, and the total number of evaluations is bounded by $\mathrm{FE}_{\max}$. 

This gives $O(\mathrm{FE}_{\max}/N)$ iterations and yields the total arithmetic cost $O\!\left(\mathrm{FE}_{\max}(\log N + D)\right)$, excluding the cost of the objective function itself. When $D$ dominates $\log N$, the total arithmetic cost reduces to $O(\mathrm{FE}_{\max}D)$.

\section{Experiments}
\label{sec:experiments}

We evaluate Myco on the CEC 2022 single-objective bound-constrained benchmark suite using functions F1 to F11 at dimensions $D=10$ and $D=20$. The experiments compare final-error performance, convergence behaviour, dimensional changes, and the contributions of cord plasticity and the community-detection backend. We first define the experimental protocol and comparator set. We then report the main comparative results and analyse the internal ablations.

\subsection{Experimental Setup}
\label{sec:experimental_setup}

Table~\ref{tab:experimental_protocol} summarises the experimental protocol used for the CEC 2022 benchmark evaluation~\cite{cec2022}. The experiments comprise 30 independent runs per algorithm and function. Myco achieved its reported results with 30 tips under an equal FE budget, while the comparator implementations used their configured population size of 50. We grouped the comparator methods by algorithmic family in Table~\ref{tab:comparator_families}. Final errors are reported by their mean and standard deviation. To facilitate reproducibility, the Python implementation of Myco is publicly available at \url{https://github.com/dehshibi/Mycelia-Search}.

\begin{table}
    \centering
    \caption{Experimental protocol for the CEC 2022 benchmark evaluation.}
    \label{tab:experimental_protocol}
        \begin{tabular}{lll}
            \hline
            \textbf{Protocol component} & \boldmath $D=10$ & \boldmath $D=20$ \\
            \hline
            Test functions & F1--F11 & F1--F11 \\
            Independent runs & 30 & 30 \\
            Maximum function evaluations & $200{,}000$ & $1{,}000{,}000$ \\
            Evaluation budget & $20{,}000D$ & $50{,}000D$ \\
            Search bounds & $[-100,100]$ & $[-100,100]$ \\
            Myco population & 30 tips & 30 tips \\
            Comparator population & 50 individuals & 50 individuals \\
            \hline
    \end{tabular}
\end{table}

\begin{table}
    \centering
    \caption{External comparator set organised by algorithm family.}
    \label{tab:comparator_families}
    \begin{tabular}{ll}
        \hline
        \textbf{Algorithm family} & \textbf{Optimisers} \\
        \hline
        Differential evolution & jSO~\cite{brest2017single}, SAP-DE~\cite{teo2005differential}, L-SHADE~\cite{tanabe2014improving}, JADE~\cite{zhang2009jade} \\
        Swarm intelligence & CLPSO~\cite{liang2006comprehensive}, ABC~\cite{karaboga2006powerful} \\
        Genetic & GA~\cite{whitley1994genetic} \\
        Mammal-inspired & GWO~\cite{mirjalili2014grey}, WOA~\cite{mirjalili2016whale} \\
        Decentralised growth & SMA~\cite{li2020slime}, MGO~\cite{zheng2024moss} \\
        \hline
    \end{tabular}
\end{table}

Following the benchmark rule, random seeds were generated deterministically for the function $f$ and run $r$ by $I_{f,r} = fR + r - R$, followed by $\sigma_{f,r} = (I_{f,r} \bmod 1000) + 1$, where $R=30$ is the number of independent runs. Convergence was recorded at 16 predefined checkpoints given by $\lfloor D^{k/5-3}\,\mathrm{FE_{\max}} \rfloor$ for $k=0,\dots,15$, where $\mathrm{FE}_{\max}$ denotes the maximum number of function evaluations.

\subsection{Results}
\label{sec:results}

Tables~\ref{tab:results_d10} and~\ref{tab:results_d20} report the final-error distributions at $D=10$ and $D=20$. The family ordering separates comparisons among methods that use different search principles. Boldface identifies the lowest mean final error for each function. Shading identifies the lowest mean within each non-differential-evolution family. The latter distinction is useful because a global row-wise winner does not, by itself, show whether the proposed network mechanism is competitive with search methods built on different information-sharing structures.

\begin{sidewaystable}[!htbp]
    \centering
    \caption{Final errors at $D=10$ over 30 independent runs, reported as mean $\pm$ standard deviation. The reported error is the difference between the obtained objective value and the known optimum of the corresponding CEC 2022 function. Algorithms are grouped by search family. Boldface identifies the lowest mean final error for each function. Light shading identifies the lowest mean within the Growth Network, Genetic, Swarm Intelligence, Mammal-inspired, and Decentralised Growth families.}
    \label{tab:results_d10}
    \renewcommand{\arraystretch}{1.08}
    \resizebox{\textheight}{!}{
        \begin{tabular}{lcccccccc
        >{\columncolor[HTML]{ECF4FF}}c 
        >{\columncolor[HTML]{ECF4FF}}c 
        >{\columncolor[HTML]{ECF4FF}}c 
        >{\columncolor[HTML]{ECF4FF}}c }
            \hline
             & \textbf{Growth Network} & \textbf{Genetic/Evolutionary} & \multicolumn{2}{c}{\textbf{Swarm Intelligence}} & \multicolumn{2}{c}{\textbf{Mammal-inspired}} & \multicolumn{2}{c}{\textbf{Decentralised Growth}} & \multicolumn{4}{c}{\cellcolor[HTML]{ECF4FF}\textbf{Differential Evolution}} \\ \cline{2-13} 
            \multirow{-2}{*}{\textbf{Function}} & Myco & GA & CLPSO & ABC & WOA & GWO & SMA & MGO & jSO & SAP-DE & L-SHADE & JADE \\ \hline
            F1 & \cellcolor[HTML]{C0C0C0}$\mathbf{0 \pm 0}$ & $3494 \pm 3337$ & $1429 \pm 486$ & $4594 \pm 1206$ & $498 \pm 1136$ & $14.5 \pm 19.7$ & $6.89\times10^{-4} \pm 3.45\times10^{-4}$ & $4900 \pm 1797$ & $\mathbf{0 \pm 0}$ & $5635 \pm 2189$ & $\mathbf{0 \pm 0}$ & $\mathbf{0 \pm 0}$ \\
            F2 & \cellcolor[HTML]{C0C0C0}$6.14 \pm 2.43$ & $7.34 \pm 11.6$ & $21 \pm 13.8$ & $8.74 \pm 0.739$ & $11.7 \pm 9.38$ & $7.37 \pm 3.14$ & $7.25 \pm 1.87$ & $60 \pm 30.1$ & $4.78 \pm 3.53$ & $\mathbf{3.19 \pm 2.98}$ & $7.14 \pm 2.6$ & $6.45 \pm 2.46$ \\
            F3 & $3.61 \pm 0.418$ & $3.2 \pm 0.374$ & $3.12 \pm 0.16$ & $3.26 \pm 0.148$ & $3.54 \pm 0.294$ & \cellcolor[HTML]{C0C0C0}$2.21 \pm 0.493$ & $2.43 \pm 0.459$ & $3.02 \pm 0.289$ & $2.71 \pm 0.475$ & $2.62 \pm 0.315$ & $2.34 \pm 0.278$ & $\mathbf{1.8 \pm 0.373}$ \\
            F4 & $178 \pm 737$ & $70.4 \pm 22.9$ & $69.9 \pm 9.05$ & $40.1 \pm 4.9$ & $1565 \pm 1073$ & \cellcolor[HTML]{C0C0C0}$26.8 \pm 11.3$ & $29.5 \pm 8.8$ & $903 \pm 390$ & $22.9 \pm 7.76$ & $22.6 \pm 4.55$ & $9.16 \pm 4.55$ & $\mathbf{7.28 \pm 3.38}$ \\
            F5 & $524 \pm 617$ & $97.6 \pm 120$ & $26 \pm 11.3$ & $8.93 \pm 3.66$ & $566 \pm 343$ & $1.54 \pm 4.44$ & \cellcolor[HTML]{C0C0C0}$0.0727 \pm 0.154$ & $158 \pm 63.6$ & $0.0966 \pm 0.172$ & $7.83 \pm 6.29$ & $\mathbf{0 \pm 0}$ & $\mathbf{0 \pm 0}$ \\
            F6 & \cellcolor[HTML]{C0C0C0}$92 \pm 143$ & $7034 \pm 5061$ & $3.08\times10^{4} \pm 8980$ & $428 \pm 268$ & $4066 \pm 4026$ & $3295 \pm 2482$ & $3184 \pm 2124$ & $1.51\times10^{4} \pm 4851$ & $3.55 \pm 1.7$ & $205 \pm 380$ & $\mathbf{0.309 \pm 0.218}$ & $0.369 \pm 0.111$ \\
            F7 & $1691 \pm 4411$ & $2837 \pm 3084$ & $4.00\times10^{4} \pm 2.55\times10^{4}$ & $2472 \pm 605$ & $3.50\times10^{4} \pm 1.86\times10^{5}$ & $3431 \pm 2826$ & \cellcolor[HTML]{C0C0C0}$1614 \pm 1054$ & $9.39\times10^{4} \pm 8.30\times10^{4}$ & $20.2 \pm 84.6$ & $240 \pm 313$ & $\mathbf{2.57 \pm 13}$ & $41.9 \pm 125$ \\
            F8 & $20.1 \pm 0.233$ & \cellcolor[HTML]{C0C0C0}$6.34 \pm 8.94$ & $20.4 \pm 0.542$ & $20.2 \pm 0.525$ & $20 \pm 0.156$ & $18.3 \pm 6.1$ & $19.3 \pm 3.59$ & $19.8 \pm 0.866$ & $12.5 \pm 9.4$ & $0.54 \pm 1.87$ & $\mathbf{0.057 \pm 0.0661}$ & $11.9 \pm 7.03$ \\
            F9 & $3.97 \pm 2.9$ & $346 \pm 639$ & $8.46\times10^{4} \pm 2.77\times10^{4}$ & \cellcolor[HTML]{C0C0C0}$\mathbf{0.9 \pm 0}$ & $4.86 \pm 2.84$ & $2.68\times10^{4} \pm 2.80\times10^{4}$ & $0.907 \pm 0.012$ & $1.62\times10^{7} \pm 1.26\times10^{7}$ & $1.7 \pm 2.06$ & $6.72 \pm 16.1$ & $\mathbf{0.9 \pm 0}$ & $\mathbf{0.9 \pm 0}$ \\
            F10 & \cellcolor[HTML]{C0C0C0}$19.8 \pm 19.3$ & $298 \pm 340$ & $1466 \pm 766$ & $25.6 \pm 11.3$ & $54.7 \pm 55$ & $670 \pm 2306$ & $26.7 \pm 19.6$ & $3.23\times10^{7} \pm 2.73\times10^{7}$ & $7.64 \pm 11.7$ & $22.9 \pm 48.9$ & $\mathbf{1.8 \pm 0}$ & $\mathbf{1.8 \pm 0}$ \\
            F11 & $4.69 \pm 1.78$ & $1.93 \pm 0.641$ & $5805 \pm 9212$ & $26.8 \pm 7.06$ & $48.8 \pm 27.8$ & $10.7 \pm 4.94$ & \cellcolor[HTML]{C0C0C0}$0.804 \pm 0.294$ & $5.63\times10^{11} \pm 5.11\times10^{11}$ & $2.03 \pm 0.845$ & $3.3 \pm 3.9$ & $0.779 \pm 0.165$ & $\mathbf{0.775 \pm 0.147}$ \\ \hline
        \end{tabular}
    }
\end{sidewaystable}

\begin{sidewaystable}[!htbp]
    \centering
    \caption{Final errors at $D=20$ over 30 independent runs, reported as mean $\pm$ standard deviation. The reported error is the difference between the obtained objective value and the known optimum of the corresponding CEC 2022 function. Algorithms are grouped by search family. Boldface identifies the lowest mean final error for each function. Light shading identifies the lowest mean within the Growth Network, Genetic, Swarm Intelligence, Mammal-inspired, and Decentralised Growth families.}
    \label{tab:results_d20}
    \renewcommand{\arraystretch}{1.08}
    \resizebox{\textheight}{!}{
        \begin{tabular}{lcccccccc
        >{\columncolor[HTML]{ECF4FF}}c 
        >{\columncolor[HTML]{ECF4FF}}c 
        >{\columncolor[HTML]{ECF4FF}}c 
        >{\columncolor[HTML]{ECF4FF}}c }
            \hline
             & \textbf{Growth Network} & \textbf{Genetic/Evolutionary} & \multicolumn{2}{c}{\textbf{Swarm Intelligence}} & \multicolumn{2}{c}{\textbf{Mammal-inspired}} & \multicolumn{2}{c}{\textbf{Decentralised Growth}} & \multicolumn{4}{c}{\cellcolor[HTML]{ECF4FF}\textbf{Differential Evolution}} \\ \cline{2-13} 
            \multirow{-2}{*}{\textbf{Function}} & Myco & GA & CLPSO & ABC & WOA & GWO & SMA & MGO & jSO & SAP-DE & L-SHADE & JADE \\ \hline
            F1 & \cellcolor[HTML]{C0C0C0}$\mathbf{0 \pm 0}$ & $2147 \pm 972$ & $3273 \pm 895$ & $2.68\times10^{4} \pm 3882$ & $7.89 \pm 42.5$ & $167 \pm 230$ & $1.66\times10^{-3} \pm 8.47\times10^{-4}$ & $2.06\times10^{4} \pm 2811$ & $\mathbf{0 \pm 0}$ & $2.31\times10^{4} \pm 7685$ & $\mathbf{0 \pm 0}$ & $\mathbf{0 \pm 0}$ \\
            F2 & \cellcolor[HTML]{C0C0C0}$\mathbf{29.9 \pm 22.7}$ & $53.6 \pm 17.4$ & $74.5 \pm 9.44$ & $79.3 \pm 4.42$ & $48.8 \pm 21.6$ & $56.7 \pm 20.9$ & $48.4 \pm 4.6$ & $227 \pm 32.1$ & $31.1 \pm 22.2$ & $32.5 \pm 13.4$ & $42.4 \pm 16.6$ & $48.8 \pm 1.04$ \\
            F3 & $8.11 \pm 0.421$ & $7.57 \pm 0.477$ & $7.5 \pm 0.206$ & $7.87 \pm 0.133$ & $8.07 \pm 0.369$ & \cellcolor[HTML]{C0C0C0}$6.02 \pm 0.571$ & $6.47 \pm 0.538$ & $7.74 \pm 0.29$ & $7.26 \pm 0.57$ & $6.94 \pm 0.36$ & $6.75 \pm 0.308$ & $\mathbf{5.81 \pm 0.525}$ \\
            F4 & $3775 \pm 3935$ & $225 \pm 60.8$ & $145 \pm 14.3$ & $233 \pm 18.6$ & $1.07\times10^{4} \pm 3544$ & $128 \pm 54.2$ & \cellcolor[HTML]{C0C0C0}$87.2 \pm 17.6$ & $6356 \pm 1595$ & $91.7 \pm 18$ & $101 \pm 15.4$ & $35.3 \pm 5.48$ & $\mathbf{29.9 \pm 9.93}$ \\
            F5 & $2654 \pm 1429$ & $477 \pm 239$ & $138 \pm 54.1$ & $1005 \pm 150$ & $2392 \pm 1107$ & $27.4 \pm 57.6$ & \cellcolor[HTML]{C0C0C0}$20.5 \pm 34.8$ & $838 \pm 225$ & $61.1 \pm 70.7$ & $467 \pm 140$ & $\mathbf{0 \pm 0}$ & $2.98\times10^{-3} \pm 0.0161$ \\
            F6 & $9.53\times10^{4} \pm 4.66\times10^{4}$ & $4.94\times10^{4} \pm 2.25\times10^{4}$ & $3.90\times10^{5} \pm 1.31\times10^{5}$ & $1.01\times10^{7} \pm 4.19\times10^{6}$ & \cellcolor[HTML]{C0C0C0}$1.84\times10^{4} \pm 1.16\times10^{4}$ & $7.20\times10^{4} \pm 2.31\times10^{4}$ & $1.47\times10^{5} \pm 2.71\times10^{4}$ & $3.35\times10^{4} \pm 6176$ & $1962 \pm 1062$ & $4.04\times10^{4} \pm 8783$ & $\mathbf{1118 \pm 534}$ & $1.23\times10^{4} \pm 1.21\times10^{4}$ \\
            F7 & \cellcolor[HTML]{C0C0C0}$2144 \pm 768$ & $5.66\times10^{4} \pm 6.28\times10^{4}$ & $1.59\times10^{5} \pm 6.25\times10^{4}$ & $1.52\times10^{6} \pm 4.50\times10^{5}$ & $1.91\times10^{5} \pm 8.20\times10^{5}$ & $1.79\times10^{5} \pm 1.91\times10^{5}$ & $9700 \pm 4999$ & $1.68\times10^{6} \pm 3.18\times10^{6}$ & $557 \pm 393$ & $8.12\times10^{4} \pm 1.21\times10^{5}$ & $\mathbf{225 \pm 316}$ & $375 \pm 428$ \\
            F8 & $20.2 \pm 0.412$ & \cellcolor[HTML]{C0C0C0}$\mathbf{19.4 \pm 3.54}$ & $21.3 \pm 0.14$ & $21.5 \pm 0.217$ & $20.3 \pm 0.397$ & $21 \pm 0.572$ & $20.1 \pm 0.0632$ & $20.1 \pm 0.0258$ & $20 \pm 0.0263$ & $20 \pm 9.32\times10^{-3}$ & $20 \pm 0.0156$ & $20.6 \pm 0.161$ \\
            F9 & \cellcolor[HTML]{C0C0C0}$6.98 \pm 3.85$ & $37.2 \pm 42.3$ & $6.93\times10^{4} \pm 2.17\times10^{4}$ & $3.28\times10^{6} \pm 9.40\times10^{5}$ & $10.4 \pm 5.36$ & $3.17\times10^{5} \pm 8.96\times10^{5}$ & $4.23 \pm 3.32$ & $3.90\times10^{7} \pm 2.54\times10^{7}$ & $6.02 \pm 4.05$ & $9.76 \pm 11.4$ & $2.77 \pm 2.23$ & $\mathbf{2.11 \pm 1.16}$ \\
            F10 & \cellcolor[HTML]{C0C0C0}$57.6 \pm 41.9$ & $356 \pm 398$ & $3253 \pm 1363$ & $1.33\times10^{6} \pm 4.37\times10^{5}$ & $122 \pm 69.9$ & $2.08\times10^{4} \pm 5.19\times10^{4}$ & $34.4 \pm 24.3$ & $5.33\times10^{7} \pm 5.12\times10^{7}$ & $75.5 \pm 46.8$ & $152 \pm 337$ & $10.1 \pm 10.3$ & $\mathbf{7.44 \pm 7.3}$ \\
            F11 & $17.8 \pm 9.52$ & \cellcolor[HTML]{C0C0C0}$\mathbf{1.54 \pm 0.0587}$ & $6080 \pm 1.03\times10^{4}$ & $6.22\times10^{10} \pm 3.39\times10^{10}$ & $132 \pm 48.8$ & $5.02\times10^{6} \pm 2.47\times10^{7}$ & $1.62 \pm 0.533$ & $7.09\times10^{12} \pm 7.76\times10^{12}$ & $16.5 \pm 9.56$ & $10.8 \pm 9.46$ & $1.92 \pm 0.722$ & $1.61 \pm 0.565$ \\ \hline
        \end{tabular}
    }
\end{sidewaystable}

The benchmark classes impose distinct search demands. F1 is a shifted and fully rotated Zakharov function with a single basin. As a unimodal function, it tests directed progress in a rotated domain. F2 contains the curved and non-separable Rosenbrock valley. F3 to F5 present increasingly difficult multimodal conditions through expanded Schaffer, non-continuous Rastrigin, and Levy structures. F6 to F8 are hybrid functions composed of transformed subcomponents, while F9-F11 are composition functions with heterogeneous local regions~\cite{cec2022}. Thus, the transition from F1 and F2 to the remaining classes shifts the principal challenge from locating a single favourable direction to maintaining search diversity while discriminating among multiple basins.

The exact convergence of Myco on F1 at both dimensions suggests that its spatial graph construction, anchor retention, and flow-driven tip motion are sufficient to reach the single-basin optimum within the present evaluation budgets. Its performance on F2 at $D=20$ further indicates that the interaction network provides each tip with a locally weighted displacement field, while anchors preserve previously favourable locations. Together, these mechanisms form a local structural memory that can maintain multiple directional signals rather than forcing the population to collapse prematurely toward a single global reference.

The basic multimodal functions place greater pressure on cross-basin exploration and late-stage coordinate refinement. Their local minima, discontinuities, and mixed components require sustained movement among competing basins. The DE variants often perform well on these functions because their mutation operators use pairwise population differences and adapt control parameters based on successful runs. This mechanism can generate broad exploratory displacements before concentrating the search around promising regions. Among these variants, L-SHADE further reduces the population size, which can shift the search toward exploitation as the run progresses~\cite{tanabe2014improving}. In contrast, Myco retains a local interaction graph. When this graph separates into spatially distinct groups, lower inter-community interaction weights reduce the influence of distant regions. This can preserve local organisation, but it may also delay the transfer of information needed to escape deceptive basins or to cross discontinuous boundaries.

The hybrid and composition functions test whether search information remains useful when several component landscapes compete across the domain. Myco can maintain competitive errors as long as the local flow field continues to identify a productive search direction. However, its performance weakens when relevant information is distributed across disconnected or competing regions of the landscape. The results suggest a trade-off in the current implementation: community-weighted local flow supports organised exploration, but it may also restrict rapid redistribution of search effort when the landscape begins to favour a distant component.

\begin{figure}
    \centering
    \includegraphics[width=0.95\textwidth]{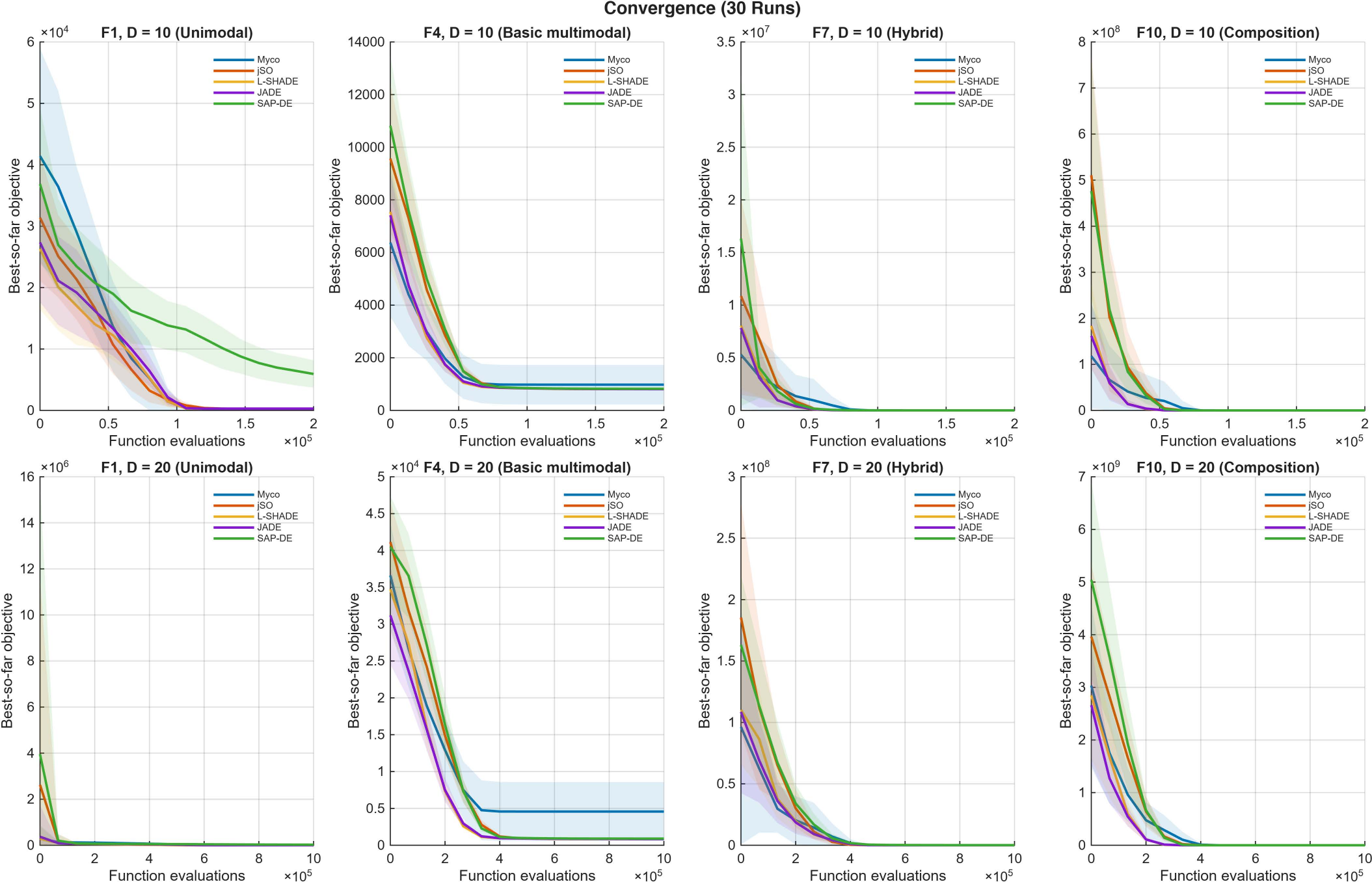}
    \caption{Mean best-so-far error trajectories over 30 runs for representative functions from the unimodal, basic multimodal, hybrid, and composition classes at $D=10$ and $D=20$. Shaded regions denote one standard deviation.}
    \label{fig:convergence}
\end{figure}

\begin{figure}
    \centering
    \includegraphics[width=0.78\textwidth]{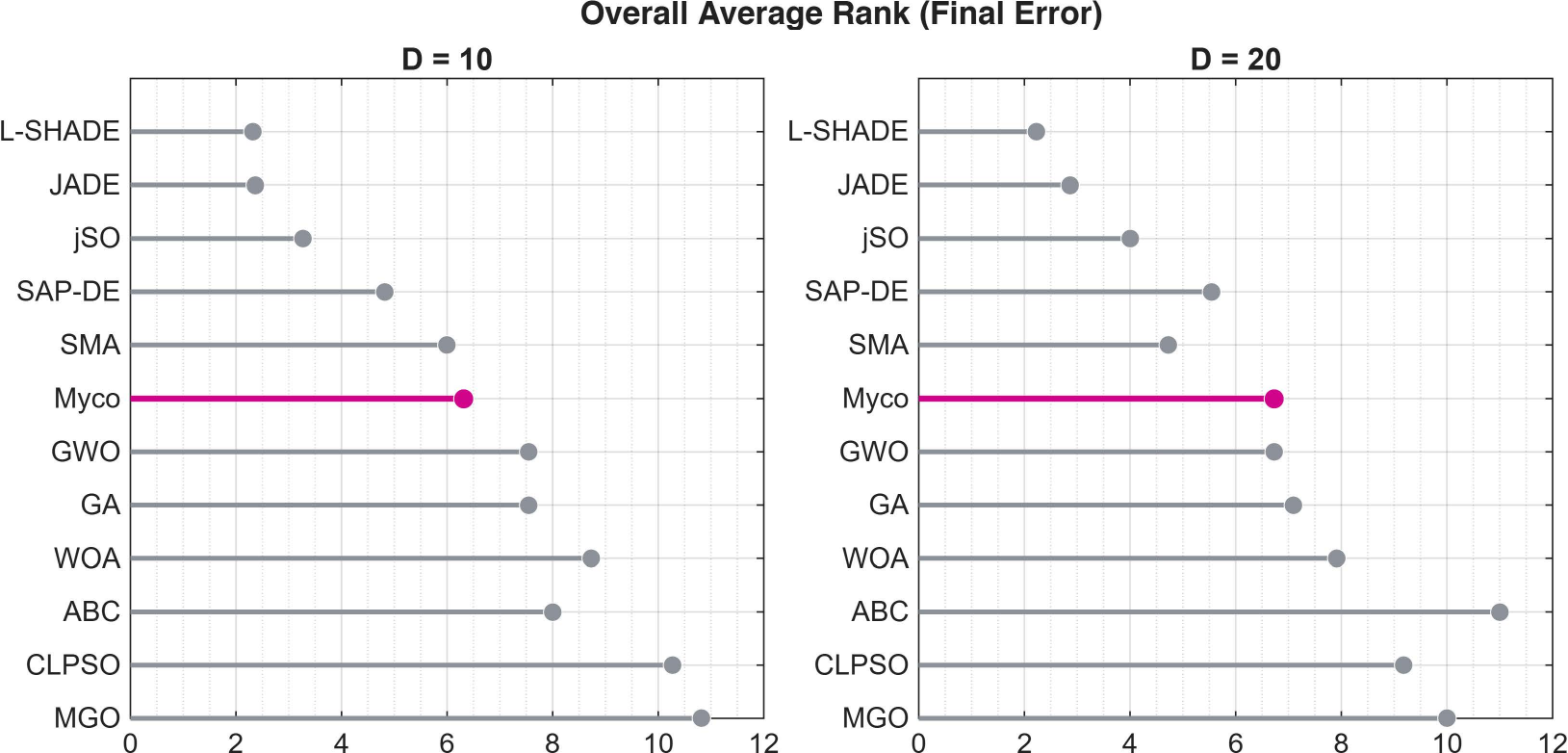}
    \caption{Average rank of each algorithm over F1 to F11 at $D=10$ and $D=20$, computed from median final error. Lower ranks indicate lower median error.}
    \label{fig:overall_rank}
\end{figure}

\begin{figure}
    \centering
    \includegraphics[width=0.5\textwidth]{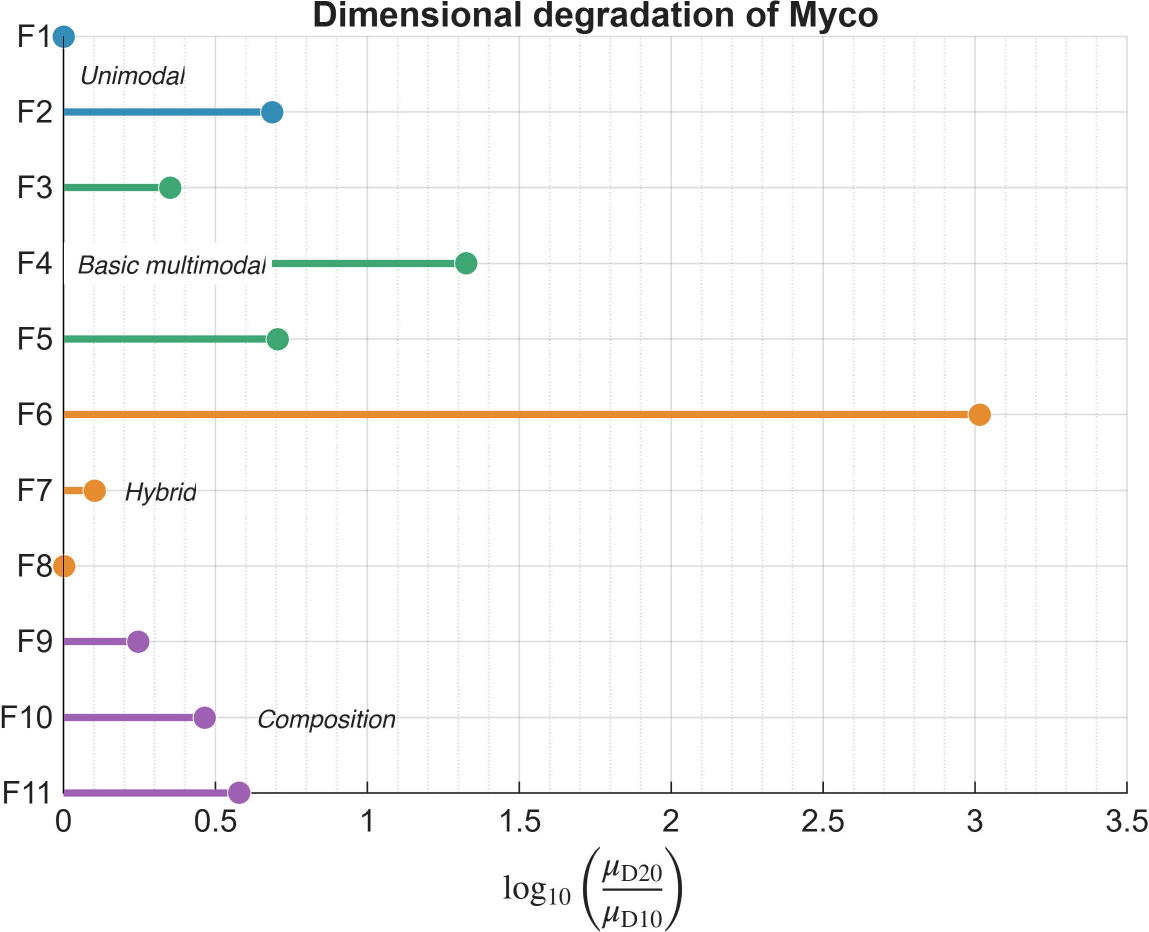}
    \caption{Change in the mean final error of Myco from $D=10$ to $D=20$ over F1 to F11, calculated as $\log_{10}(\mu_{D20}/\mu_{D10})$. Positive values indicate an increase in mean error at $D=20$. Functions with zero mean error in both dimensions are assigned a value of zero.}
    \label{fig:dimensional_change}
\end{figure}

Figure~\ref{fig:convergence} illustrates the temporal pattern of this distinction. While Figure~\ref{fig:overall_rank} provides a median-based summary that is less sensitive to extreme runs, Figure~\ref{fig:dimensional_change} isolates the change in mean final error, showing that increasing the dimension alters the final-error profile of Myco in a consistent direction.

\subsection{Ablation Study}
\label{sec:ablation_study}

We examine the respective contributions of adaptive cord conductance and the community-detection backend (i.e., Louvain and Greedy Modularity) within two ablation variants. Myco (Louvain) preserves the Louvain partition and all remaining components of the Plasticity, but disables the conductance update on tip-to-tip edges. Myco (Greedy) instead replaces the Louvain backend with Greedy Modularity optimisation. These comparisons isolate distinct aspects of the search graph: the former evaluates whether edge strengths should adapt to the induced flow, whereas the latter examines whether the resulting flow field benefits from a non-trivial mesoscale partition.

\subsubsection{Cord Plasticity}
\label{sec:cord_plasticity_ablation}

Table~\ref{tab:plasticity_ablation} compares the Louvain and Plasticity variants using paired mean and standard-deviation values. The logarithmic ratio quantifies both the direction and magnitude of each paired difference, with negative values indicating improved performance under Plasticity.

\begin{table}
    \centering
    \caption{Effect of cord plasticity relative to the Louvain baseline across 30 independent runs. Values are reported as mean $\pm$ standard deviation. The final columns provide the relative effect size, computed as $\log_{10}(\mu_{\mathrm{Plasticity}}/\mu_{\mathrm{Louvain}})$, where negative values indicate lower mean error under Plasticity and positive values indicate lower mean error under Louvain. Empty entries report cases where one or both means are zero.}
    \label{tab:plasticity_ablation}
    \resizebox{0.97\textwidth}{!}{%
    \begin{tabular}{lccc|ccc}
        \hline
        \multirow{2}{*}{\textbf{Function}} & \multicolumn{3}{c|}{\boldmath $D=10$} & \multicolumn{3}{c}{\boldmath $D=20$} \\ \cline{2-7} 
         & \textbf{Louvain} & \textbf{Plasticity} & $\log_{10}\left(\frac{\mu_{\mathrm{Plasticity}}}{\mu_{\mathrm{Louvain}}}\right)$ & \textbf{Louvain} & \textbf{Plasticity} & $\log_{10}\left(\frac{\mu_{\mathrm{Plasticity}}}{\mu_{\mathrm{Louvain}}}\right)$ \\ \hline
        F1 & $0 \pm 0$ & $0 \pm 0$ & -- & $4.95\times10^{-6} \pm 1.47\times10^{-5}$ & $0 \pm 0$ & -- \\
        F2 & $5.67 \pm 2.57$ & $6.14 \pm 2.43$ & $0.034$ & $25.6 \pm 23.7$ & $29.9 \pm 22.7$ & $0.066$ \\
        F3 & $3.52 \pm 0.382$ & $3.61 \pm 0.418$ & $0.011$ & $8.23 \pm 0.500$ & $8.11 \pm 0.421$ & $-0.007$ \\
        F4 & $727 \pm 1341$ & $178 \pm 737$ & $-0.610$ & $6848 \pm 5127$ & $3775 \pm 3935$ & $-0.259$ \\
        F5 & $551 \pm 709$ & $524 \pm 617$ & $-0.021$ & $3291 \pm 1698$ & $2654 \pm 1429$ & $-0.093$ \\
        F6 & $865 \pm 2305$ & $92.0 \pm 143$ & $-0.973$ & $9.12\times10^{4} \pm 3.94\times10^{4}$ & $9.53\times10^{4} \pm 4.66\times10^{4}$ & $0.019$ \\
        F7 & $3110 \pm 5507$ & $1691 \pm 4411$ & $-0.265$ & $3663 \pm 4335$ & $2144 \pm 768$ & $-0.233$ \\
        F8 & $20.10 \pm 0.166$ & $20.09 \pm 0.233$ & $-0.000$ & $20.21 \pm 0.398$ & $20.18 \pm 0.412$ & $-0.001$ \\
        F9 & $3.17 \pm 2.99$ & $3.97 \pm 2.90$ & $0.097$ & $7.79 \pm 4.41$ & $6.98 \pm 3.85$ & $-0.048$ \\
        F10 & $16.3 \pm 21.7$ & $19.8 \pm 19.3$ & $0.083$ & $45.4 \pm 23.7$ & $57.6 \pm 41.9$ & $0.103$ \\
        F11 & $4.68 \pm 3.37$ & $4.69 \pm 1.78$ & $0.001$ & $30.4 \pm 14.5$ & $17.8 \pm 9.52$ & $-0.233$ \\ \hline
    \end{tabular}%
    }
\end{table}

At $D=10$, Plasticity achieves its largest reductions on F4 and F6, with corresponding decreases in standard deviation. In the Louvain backend, existing edges are weighted according to distance, conductance, potential difference, and community membership, whereas tip-to-tip conductances remain fixed throughout the search. By contrast, Plasticity changes this local graph after the flow field has been computed. Such changes reinforce tip-to-tip edges that are aligned with the current flow while attenuating unsupported edges. The observed reduction in both location and dispersion suggests that adaptive conductance suppresses transient local directions that would otherwise retain excessive influence.

The effect becomes pronounced at $D=20$. Plasticity reduces the mean error on F1, F3, F4, F5, F7, F8, F9, and F11. The paired reductions on F4 and F7 are associated with substantially lower standard deviations. This pattern suggests that the conductance update helps preserve locally persistent search directions when distance-based neighbourhoods become less informative in higher-dimensional landscapes. However, the benefit is not universal. Louvain remains preferable on F2, F6, and F10, indicating that conductance decay can, in some landscapes, weaken alternative local directions before their search value becomes evident.

\begin{figure}
    \centering
    \includegraphics[width=0.75\textwidth]{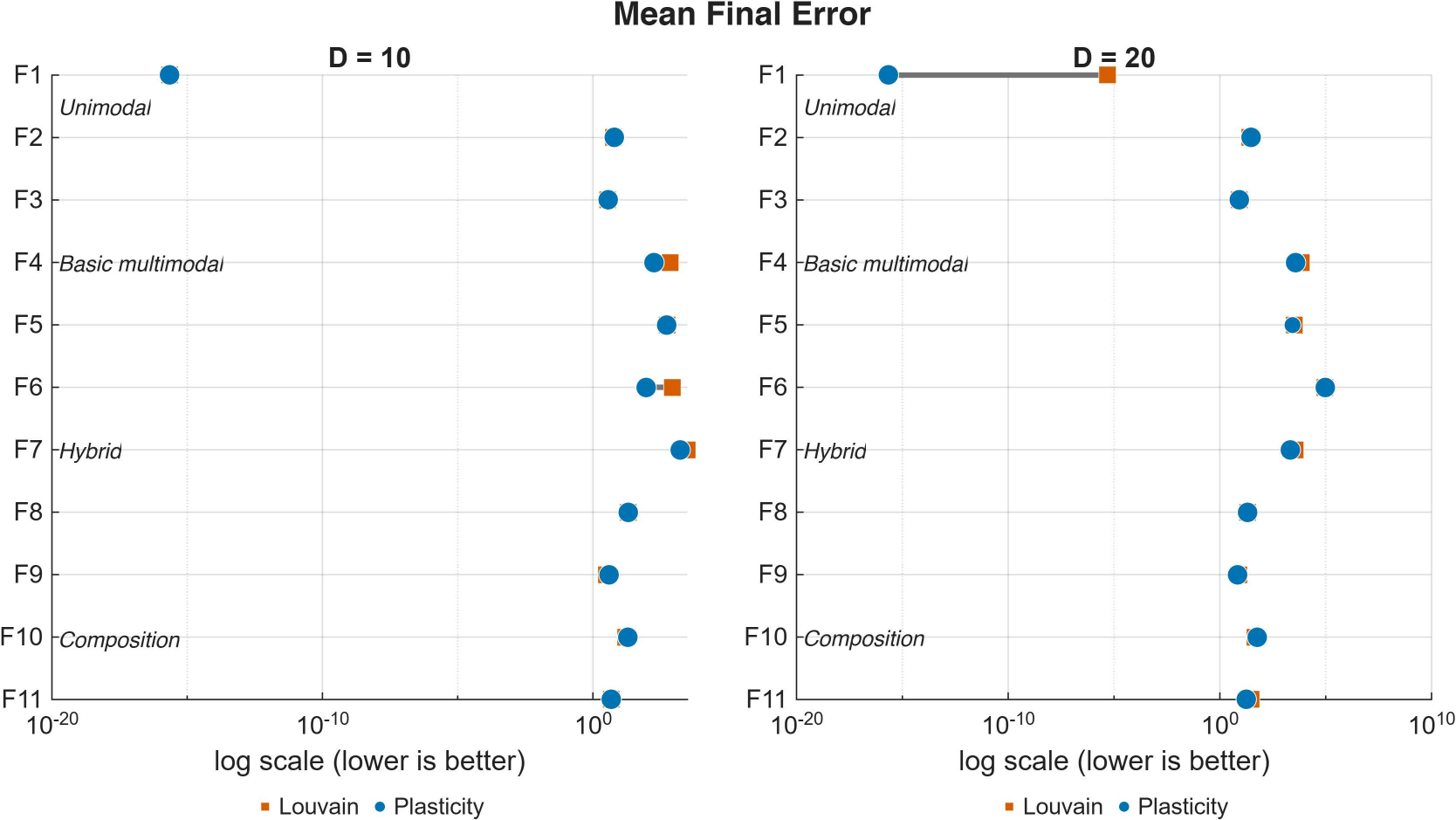}
    \caption{Mean final errors of Myco (Louvain) and Myco (Plasticity) over F1 to F11 at $D=10$ and $D=20$. Lower values are better. The horizontal axis is logarithmic.}
    \label{fig:plasticity_error}
\end{figure}

\begin{figure}
    \centering
    \includegraphics[width=0.74\textwidth]{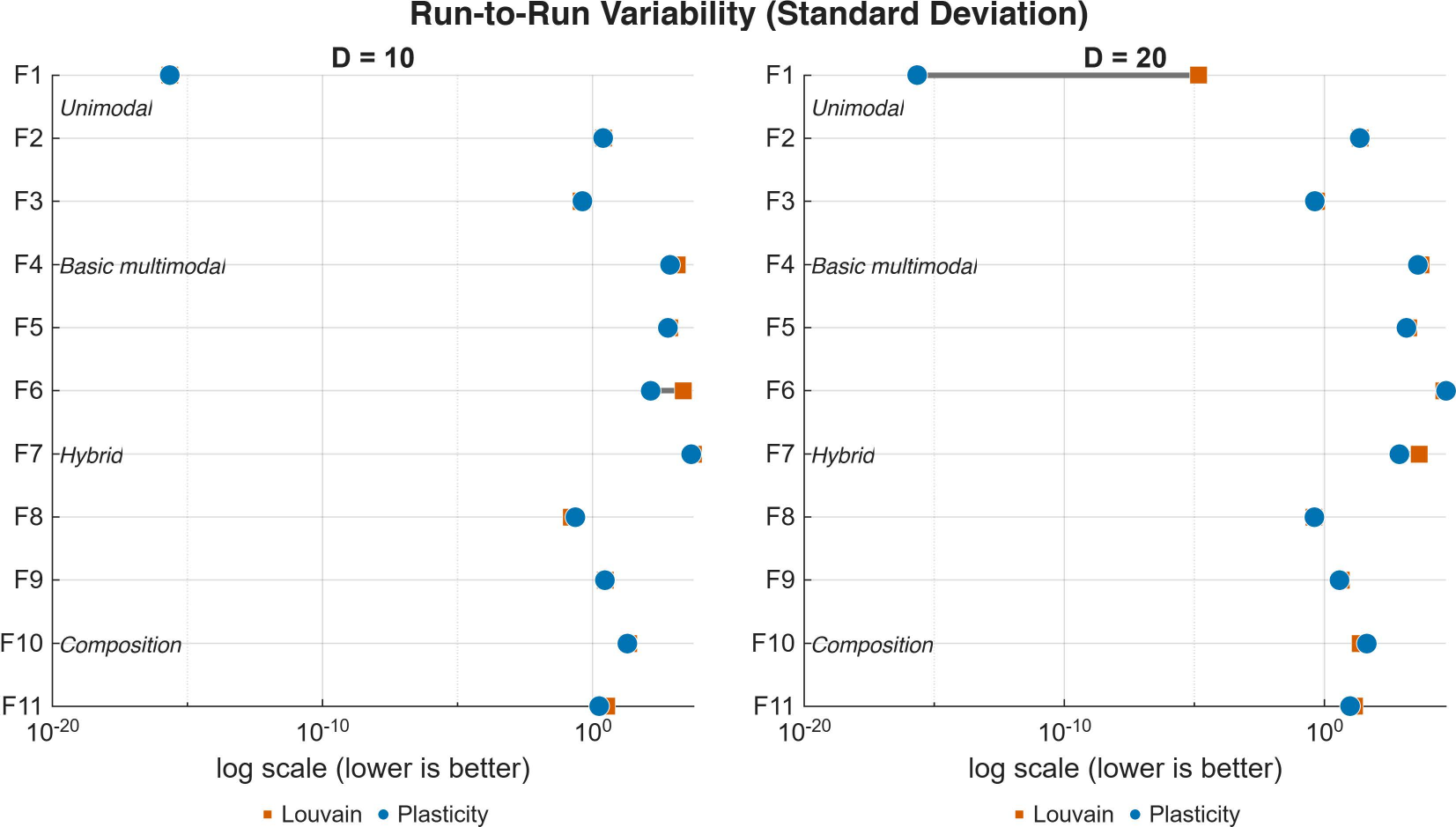}
    \caption{Standard deviations of final error for Myco (Louvain) and Myco (Plasticity) over 30 independent runs. Lower values indicate lower run-to-run variability. The horizontal axis is logarithmic.}
    \label{fig:plasticity_variability}
\end{figure}

Figure~\ref{fig:plasticity_error} shows the paired final-error outcomes, whereas Figure~\ref{fig:plasticity_variability} demonstrates the corresponding run-to-run variability. Table~\ref{tab:plasticity_ablation} complements these visual summaries by providing the exact relative changes that distinguish modest paired differences from order-of-magnitude reductions.

\subsubsection{Community Backend Sensitivity}
\label{sec:community_backend_sensitivity}

\begin{figure}
    \centering
    \begin{subfigure}
      \centering
      \includegraphics[width=0.40\linewidth]{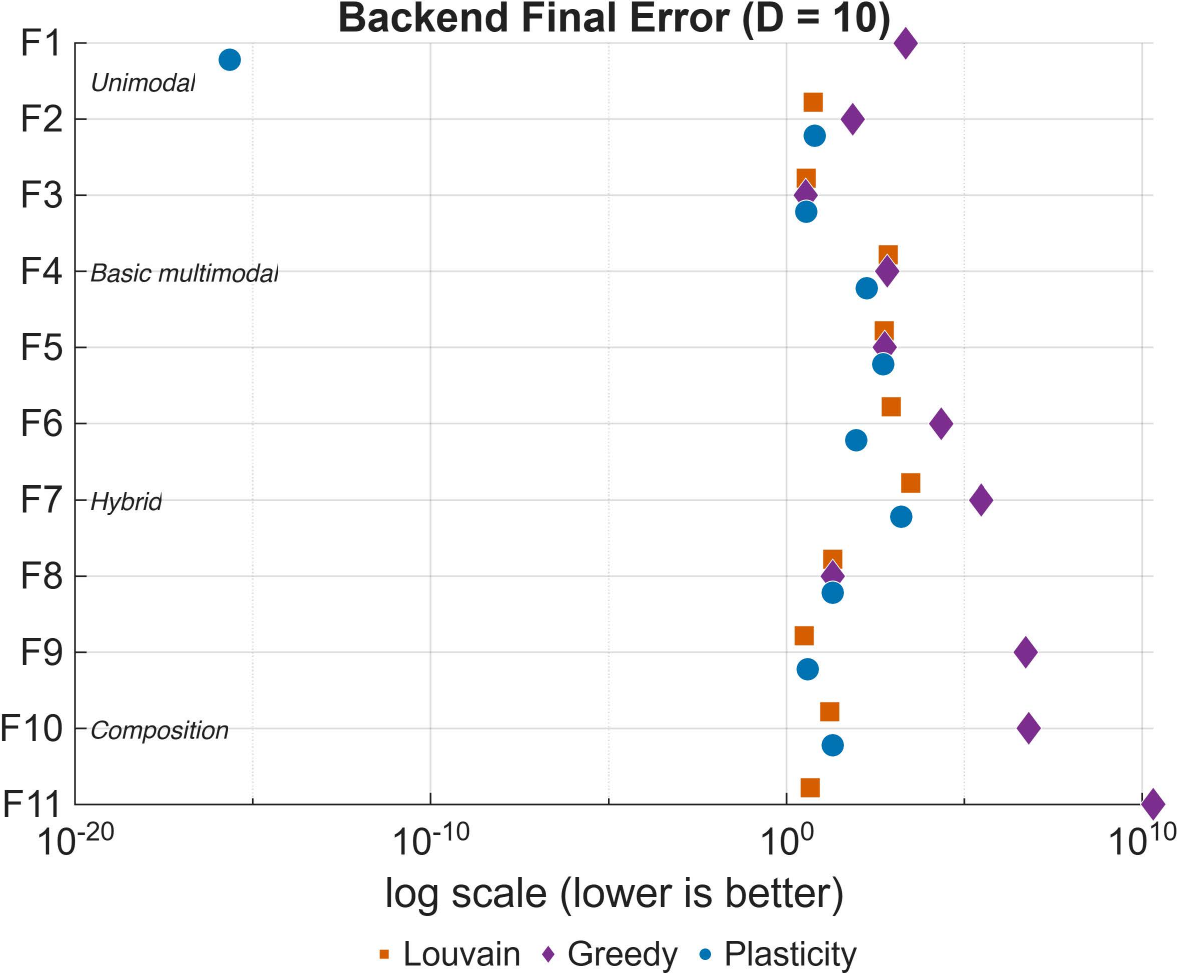}
      \label{fig:backend_error_D10}
    \end{subfigure}%
    \begin{subfigure}
      \centering
      \includegraphics[width=0.40\linewidth]{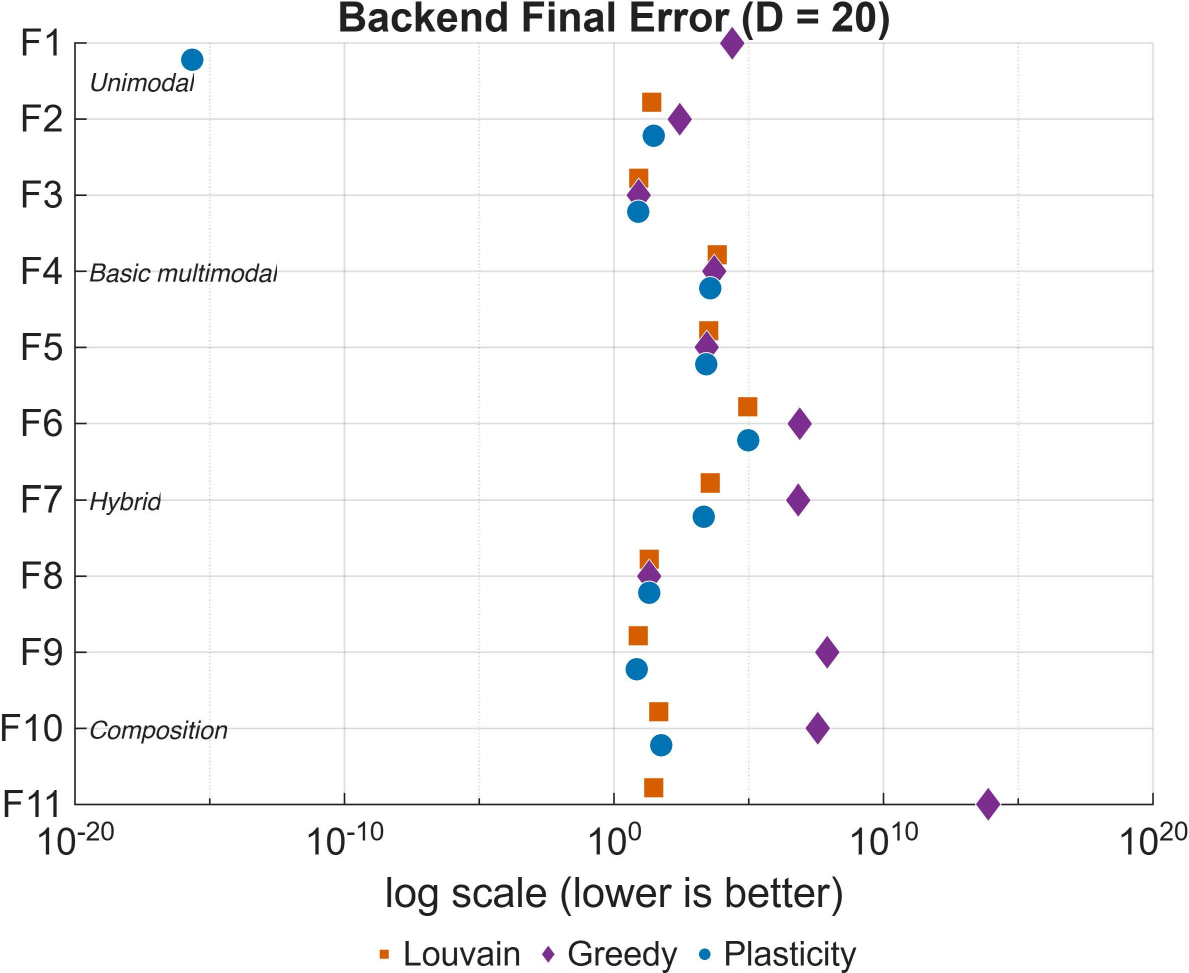}
      \label{fig:backend_error_D20}
    \end{subfigure}
    \caption{Mean final errors of the Louvain, Greedy Modularity, and Plasticity implementations over F1 to F11 at $D=10$ and $D=20$. Lower values are better. The horizontal axis is logarithmic.}
    \label{fig:backend_error}
\end{figure}

Table~\ref{tab:greedy_diagnostic} and Figure~\ref{fig:backend_error} report the sensitivity of the flow field to the choice of community detection backend. Across all reported functions and dimensions, the Greedy Modularity identifies a single raw community and a single retained community. Consequently, the distinction between intra-community and inter-community interactions is lost, and all present edges receive the same community factor.

\begin{table}
    \centering
    \caption{Greedy Modularity diagnostic for functions with large divergence between the mean and median final errors. Each entry reports 30 independent runs. The ratio $\mu/\tilde{x}$ indicates the extent to which rare high-error runs influence the mean.}
    \label{tab:greedy_diagnostic}
    \begin{tabular}{ccccc}
        \hline
        Dimension & Function & Mean ($\mu$) & Median ($\tilde{x}$) & $\mu/\tilde{x}$ \\ \hline
        $D=10$ & F1 & $2217$ & $3.15\times10^{-2}$ & $7.05\times10^{4}$ \\
        $D=10$ & F7 & $2.99\times10^{5}$ & $6239.5$ & $48.0$ \\
        $D=10$ & F9 & $5.31\times10^{6}$ & $6.07$ & $8.76\times10^{5}$ \\
        $D=10$ & F10 & $6.44\times10^{6}$ & $1146.8$ & $5.62\times10^{3}$ \\
        $D=10$ & F11 & $2.09\times10^{10}$ & $4.16$ & $5.01\times10^{9}$ \\
        $D=20$ & F6 & $7.81\times10^{6}$ & $1.03\times10^{5}$ & $75.9$ \\
        $D=20$ & F7 & $6.83\times10^{6}$ & $1.83\times10^{5}$ & $37.4$ \\
        $D=20$ & F10 & $3.65\times10^{7}$ & $3.70\times10^{6}$ & $9.86$ \\
        $D=20$ & F11 & $7.72\times10^{13}$ & $48.5$ & $1.59\times10^{12}$ \\ \hline
        \end{tabular}
\end{table}

The Greedy results contain extreme mean-to-median divergence for selected functions. At $D=10$, this behaviour is most evident on F1, F9, F10, and F11, whereas at $D=20$ it is most pronounced on F6, F7, F10, and F11. The low medians observed in several cases indicate that the one-community configuration can occasionally reach favourable regions of the search space. However, the substantially larger means suggest that a subset of runs enters poor regions from which recovery is not achieved within the available evaluation budget.

Louvain restores a partitioned flow field in which cross-community messages are weighted less strongly than within-community messages. Plasticity subsequently adjusts the relative strength of the remaining tip-to-tip interactions. These mechanisms therefore operate at distinct graph scales: Louvain regulates communication among local groups, whereas Plasticity controls the persistence of search directions within the current local graph.

\section{Conclusion}
\label{sec:conclusion}

In this study, we introduced Mycelial Search (Myco) to frame continuous optimisation as a search over an evolving local interaction graph. Candidate solutions are represented as active tips, while a bounded set of anchors preserves historically favourable positions. The graph itself is reconstructed at each iteration. We used the Louvain community detection algorithm to control the strength of information exchange within and across communities. We introduced cord plasticity, a rule that reinforces tip-to-tip connections aligned with the local flow and attenuates unsupported connections.

We evaluated Myco on the CEC 2022 single-objective bound-constrained benchmark suite, compared it with the established metaheuristic methods, and examined the roles of cord plasticity and the community detection backend in an ablation study. The ablation study results demonstrated that these mechanisms control different aspects of the search. Cord plasticity operates at the edge level, reinforcing or decaying individual tip-to-tip connections based on their alignment with the local flow. The community detection backend operates at the partition level, determining the range over which information is exchanged. 

Myco makes the interaction structure among candidate solutions explicit and subject to analysis during the search. The local graph is not only a mechanism for selecting neighbours. Its community partition and tip-to-tip edge conductances form part of the search state and affect how information is passed between candidate solutions. This formulation makes local organisation a component of continuous optimisation that can be defined, examined, and modified.


\section*{Declaration of generative AI and AI-assisted technologies in the manuscript preparation process}

During the preparation of this manuscript, the author used Grammarly to review spelling and grammar. The author reviewed and edited the resulting suggestions as needed and takes full responsibility for the content of the published article.


\bibliographystyle{cas-model2-names}

\bibliography{references}

\end{document}